\documentclass[conference]{IEEEtran}
\IEEEoverridecommandlockouts
\usepackage{cite}
\usepackage{amsmath,amssymb,amsfonts}
\usepackage{algorithmic}
\usepackage{graphicx}
\usepackage{textcomp}
\usepackage{xcolor}
\usepackage[numbers]{natbib}
\usepackage{soul}
\usepackage[T1]{fontenc}

\def\BibTeX{{\rm B\kern-.05em{\sc i\kern-.025em b}\kern-.08em
    T\kern-.1667em\lower.7ex\hbox{E}\kern-.125emX}}

\makeatletter
\newcommand{\linebreakand}{%
  \end{@IEEEauthorhalign}
  \hfill\mbox{}\par
  \mbox{}\hfill\begin{@IEEEauthorhalign}
}
\makeatother

\begin{document}

\title{Embedding Contextual Information through Reward Shaping in Multi-Agent Learning: A Case Study from Google Football}

\author{Chaoyi Gu, Varuna De Silva, Corentin Artaud, Rafael Pina}

\IEEEpubid{\begin{minipage}{\textwidth}\ \\[12pt]
  979-8-3503-3337-4/23/\$31.00 ©2023 IEEE \\ 
  DOI: 10.1109/ICPRS58416.2023.10179030 \\
  Link: https://ieeexplore.ieee.org/document/10179030
\end{minipage}} 


\maketitle
\IEEEpubidadjcol

\begin{abstract}
Artificial Intelligence has been used to help human
complete difficult tasks in complicated environments by providing
optimized strategies for decision-making or replacing the manual
labour. In environments including multiple agents, such as
football, the most common methods to train agents are Imitation
Learning and Multi-Agent Reinforcement Learning (MARL).
However, the agents trained by Imitation Learning cannot outperform the expert demonstrator, which makes humans hardly
get new insights from the learnt policy. Besides, MARL is prone
to the credit assignment problem. In environments with sparse
reward signal, this method can be inefficient. The objective
of our research is to create a novel reward shaping method
by embedding contextual information in reward function to
solve the aforementioned challenges. We demonstrate this in the
Google Research Football (GRF) environment. We quantify the
contextual information extracted from game state observation
and use this quantification together with original sparse reward
to create the shaped reward. The experiment results in the GRF
environment prove that our reward shaping method is a useful
addition to state-of-the-art MARL algorithms for training agents
in environments with sparse reward signal.
\end{abstract}


\section{Introduction}
Training AI to complete human tasks is beneficial in lots of aspects of human life, such as engineering and medical surgery. One of the key challenges is how to train AI efficiently to complete a sequence of difficult tasks to achieve the best outcome. Moreover, if it is a multi-agent system, the agents are trained to not just complete each task individually but cooperate with each other as well. To overcome the challenges, Multi-Agent Reinforcement Learning (MARL) and Imitation Learning are two main approaches made in previous studies.

MARL focuses on studying how multiple agents reach a common goal through learning, cooperating, and interacting with each other in a shared environment \cite{tan1993multi}. There are two main obstacles when using MARL to train agents accomplishing tasks. The first is that in some environments such as football, since only global reward signal is available, the agents can hardly learn the impact of individual actions on the global outcome \cite{juozapaitis2019explainable}. To solve this problem, the state-of-the-art algorithms introduced are Value Decomposition Networks (VDN) and QMIX. VDN is capable of decomposing the global reward signal to each individual agent through back-propagating the total Q gradient through deep neural networks representing the individual component value functions \cite{sunehag2017value}. Unlike VDN summing Q-values for each individual, it has a deep neural network in which parameters are learned to approximate the joint action Q-value \cite{rashid2020monotonic}. Compared to VDN, QMIX enables the shared reward factorization and is expected to make each individual agent perform better with more precise feedback. Nevertheless, QMIX only accepts monotonic factorization and requires global status information while learning, which limits its usages in complex environments \cite{rashid2020monotonic}. One of the major problems which VDN and QMIX cannot solve is the credit assignment problem \cite{zhou2020learning}. In environments like football, the naturally delayed reward can be determined by a sequence of agents' actions while agents will hardly learn efficiently without dispensed credit for each positive action or penalty for every negative action \cite{riedmiller2018learning}. Reward shaping is regarded as one of the most efficient methods to tackle this problem \cite{zou2019reward}. Specifically, to design a good reward shaping function, rich domain knowledge is necessary for different contexts within the environment \cite{hu2020learning}. Nonetheless, the potential subjectivity of the knowledge and the inappropriate way of implementing the knowledge make reward shaping a challenging task \cite{memarian2021self}.

Imitation Learning (IL), in comparison to MARL, makes agents learn how to complete difficult tasks without reward signal \cite{hussein2017imitation}. The aim of IL is to train a policy to do decision-making utilizing existing demonstrations. The advantage of IL in training policies in multi-agent systems is that neither the reward decomposition nor the credit assignment problem need to be considered. In complex environments when the reward function is difficult to specify, IL can be a good alternative to MARL \cite{reddy2019sqil}. To train the agents using IL successfully, the expert demonstration is crucial. It is the supervised signal for guiding the agents in policy learning. When interacting with the environment, the only feedback agents can get is the quantification of mismatch between their actions against the actions taken by the expert demonstrator and the deep neural networks are utilized to find the parameters for optimizing the imitation performance of policy \cite{le2017coordinated}. It indicates that without a good expert demonstration, the policy can hardly be learnt. In some environments, demonstrations are expensive or even impossible to obtain, which limits the usage of IL. Moreover even in environments where the expert demonstrations are obtainable, there is still a limitation for IL. For instance, in simulated football environments, the tracking and event data from professional matches can be used as demonstrator for training AI footballers. However, the performance of agents will always be limited by the quality of the teams learnt from, which indicates that coaching staff can hardly gain any novel insights according to the AI performance. In areas where humans expect to gain new understanding from innovative actions of AI, IL can hardly meet the requirement \cite{hussein2017imitation}.

In this research, we propose a novel reward shaping method in MARL to overcome the aforementioned challenges and test it in Google Research Football (GRF), a simulated football environment for experimenting Reinforcement Learning algorithms. With the cutting-edge models in football, we extract patterns from raw game state observations at every learning step and implement the quantified extraction into the reward function so that the agents can get informative and immediate feedback while learning. Experiment results show that our reward shaping method is an important addition to state-of-the-art algorithms for training agents in Multi-Agent Learning.

The contributions of this paper are:
\begin{itemize}
\item We propose a novel reward shaping method for multi-agent learning.

\item The effectiveness of our reward shaping method in multi-agent learning in Google Football reveals the potential of utilizing a similar framework in other multi-agent environments.
\end{itemize}
\section{Background}

\subsection{Football Related Models}

\subsubsection{Pitch Control}
The Pitch Control function (PCF) is a function used to calculate the probability that each side of the team will control the ball at given location of the football pitch at a given game state \cite{spearman2017physics}. To compute the possibility of controlling the ball from each team considering the contextual factors, the author trains a passing probability model. The model is used to predict the successful rate for a pass with players' and ball's positions, speed and facing directions provided. As shown in Eq.~\eqref{pcm1}, each pass is converted to a Bernoulli trial in which $k$ represents the outcome of the pass (success: 1, failure: 0). Whether the pass will be successful is decided by an input $x$ and two trained parameters \(\sigma\) and \(\lambda\). 
\begin{equation}
        \label{pcm1}
        p\left(k\middle|\sigma,\ \lambda,x\right)=
        \begin{cases}
            1-p\ for\ k=0\\
            p\ \ \ \ \ \ \ \ for\ k=1
        \end{cases}
\end{equation}%
To objective of the model training is to have a group of parameters which can maximize the likelihood shown in Eq.~\eqref{pcm2}. 
\begin{equation}
    \label{pcm2}
    \mathcal{L}\left(\sigma,\ \lambda\middle|\ k,x\right)=\ P(k|\sigma,\ \lambda,x)
\end{equation}%
Then, the author uses the trained passing probability model to compute the probability of a pass being successful given the location of passing target. As a result, in an imaginary location on the pitch, the probability of each team controlling the ball can be computed by this model. Furthermore, the whole pitch can be converted to a real valued scalar field to quantify the space controlled by each team with the PCF.

\subsubsection{Expected Possession Value}
Expected Possession Value (EPV) is a model used to quantify the value of ball possession at a given game state with contextual factors including the ball position and the football match state such as open play (Eq.~\eqref{epv1}) \cite{rudd_2011}.
\begin{equation}
    \label{epv1}
    P_{poss}\left(G\middle|situation\right)=\ P_{poss}(G|ball,\ match\ state)
\end{equation}%
In this study, each possession sequence in a football game is viewed as a Markov Process in which the possible next state is only depending on the current state. Compared to the traditional possession evaluation metric, Xg model \cite{rathke2017examination}, the EPV model computes the value of possession not just based on the current game state, but on the impact of the current state on the future as well, which makes EPV a better tool for evaluating the performance from each team.

\subsection{Multi-Agent Reinforcement Learning}
Multi-Agent Reinforcement Learning (MARL) is a subset of Reinforcement Learning which studies how agents develop policies for taking actions to maximize the cumulative rewards through trial and error \cite{kaelbling1996reinforcement}. Unlike single-agent Reinforcement Learning, MARL studies how multiple agents interact with each other in a common environment. MARL makes agents learn cooperative behaviors to maximize the group reward \cite{tan1993multi}.
\subsubsection{Proximal Policy Optimization}
Proximal Policy Optimization (PPO) is a method widely used in Reinforcement Learning, including actor model interacting with environment and critic model evaluating the value of actions made by the actor model \cite{yu2021surprising}. This algorithm can be unstable in Multi-Agent tasks since all of the agents can only have access to the group feedback and don’t know exactly how their actions influence the group performance \cite{guo2020joint}. Based on previous studies, when using PPO, one agent can do most jobs while others just standby doing nothing in certain episodes of learning \cite{guo2020joint}. Besides, during the training in some episodes of roll-out the agents can receive positive feedback even when they make wrong decisions, and their teammates improve the team performance \cite{bai2021improved}. Both above scenarios can make the learning go back and forth and finally fail.

\subsubsection{Value Decomposition Networks}
\begin{figure}[b]
    \centering
    \includegraphics[width=180pt]{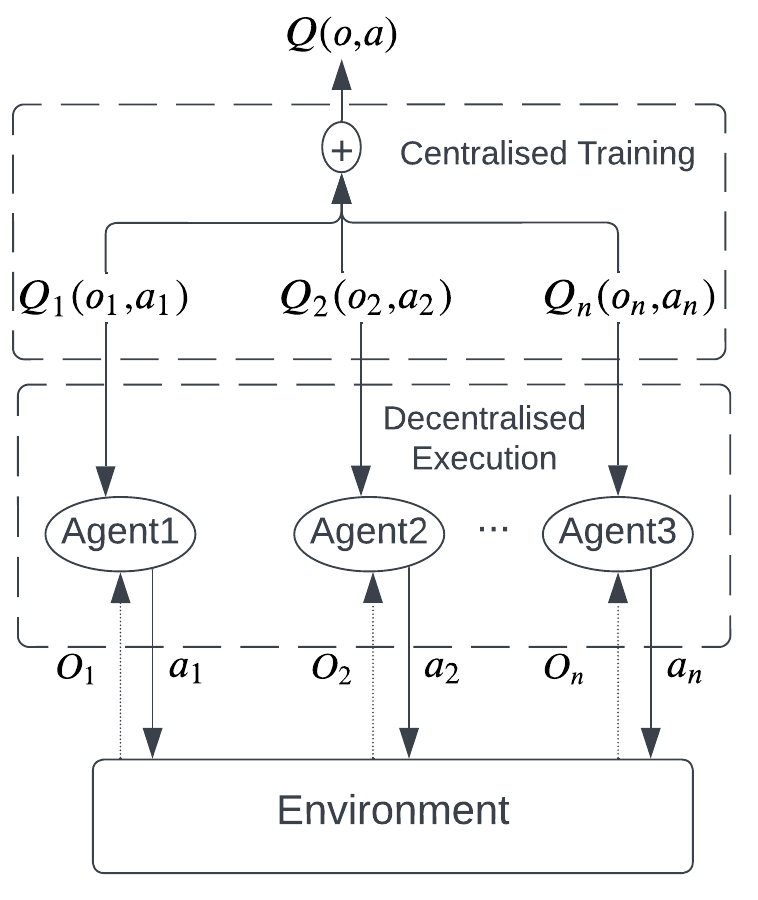}
    \caption{Architecture of Valued Decomposition Network}
    \label{vdn1}
\end{figure}

Value Decomposition Networks (VDN) is a newly introduced algorithm to solve MARL problems \cite{sunehag2017value}. Unlike the previous method, it provides customized feedback to each individual agent in terms of decomposing the group feedback. In specific, the decomposition network tries to learn an optimal value decomposition from the global reward signal thorough back-propagating the total Q gradient via deep neural networks representing the individual component value functions. The addictive value decomposition aims to avoid the spurious reward signals which emerge in every independent agent. Moreover, the value function learned by each individual depends merely on local observations. This method enables the learning set up in a centralised fashion during training while every agent can be deployed separately.
As shown in Fig.~\ref{vdn1}, value decomposition individual architecture shows how local observations enter the networks, individual values for agents are summed to a joint Q-function for training, and actions are produced independently from the individual outputs through decentralised execution.

\subsection{Google Research Football Environment}
\begin{figure}[t]
    \centering
    \includegraphics[width=200pt]{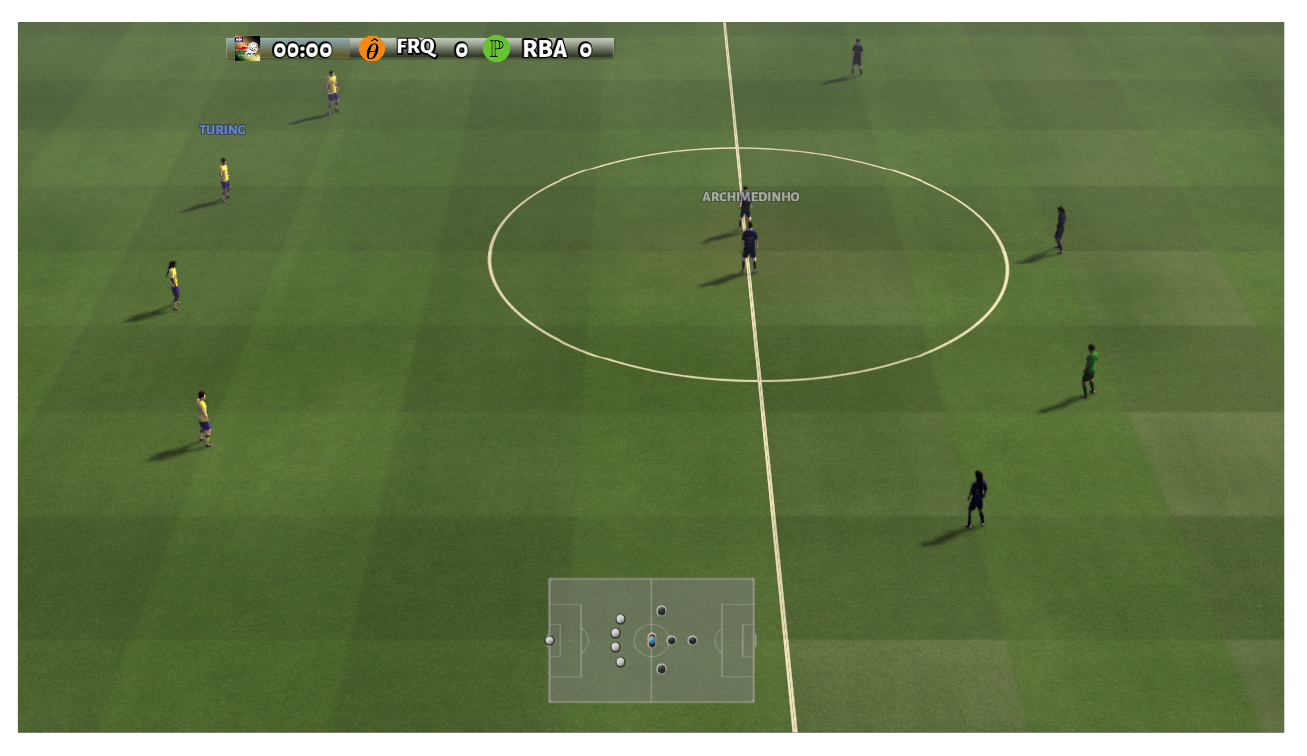}
    \caption{Google Research Football}
    \label{gfootball}
\end{figure}

Google Research Football (GRF), also known as Google Football is a virtual environment for testing Reinforcement Learning \cite{kurach2020google}. As shown in Fig.~\ref{gfootball}, it is a physics-based 3D simulator reproducing the real football game and allows algorithms to control agents playing football. The environment supports not only the single-agent experiment but multi-agent learning as well. Algorithms can be used to control one or several players and the other players within the game will be controlled by the built-in hard-coded AI. Three difficulty levels (hard:0.95, medium:0.6, easy:0.05) are provided to change the built-in AI's ability. Moreover, the environment allows customization of the game scenario so that the environment can be modified based on the algorithms required to test.

\begin{figure}[b]
    \centering
    \includegraphics[width=190pt]{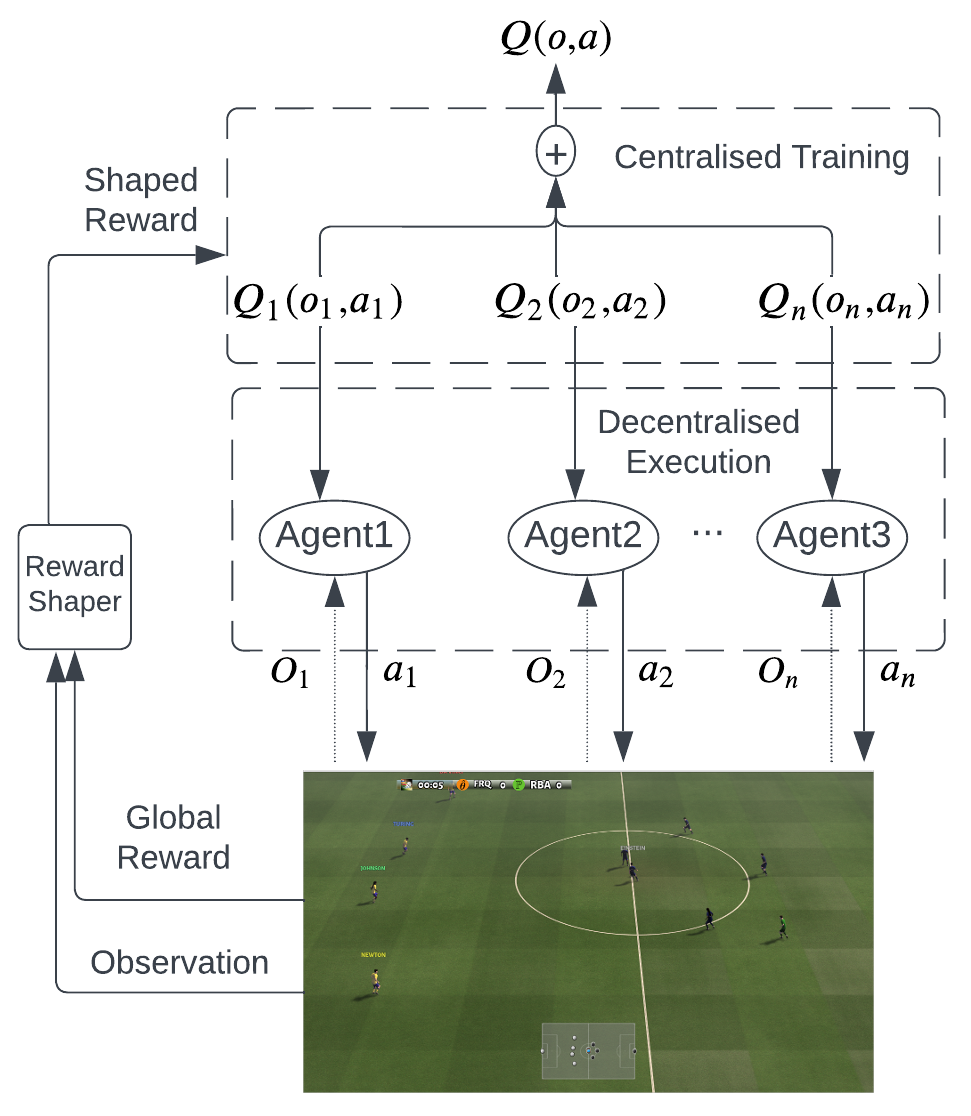}
    \caption{Architecture of learning framework}
    \label{vdn2}
\end{figure}

\section{Methodology}
\subsection{Learning Framework}
To reach the research objective which is to train a group of agents to accomplish difficult tasks in an environment with only sparse reward available, we use VDN as the learning algorithm and propose a novel learning framework to improve the agents' performance, despite the reward sparsity problem. Afterwards, the learning framework is tested in the GRF environment. As shown in Fig.~\ref{vdn2}, when each agent interacts with the environment, they receive a shaped reward. The primary reward is the goal difference between two sides. In addition to that, we add extra reward features to help them learn better. After the game state observation is extracted from the environment, the pattern within the observed game state is detected and evaluated utilizing domain knowledge. Then the pattern quantification is added to the primary reward to give each agent more insightful feedback.

\subsection{Quantification of Patterns Detected in Raw Observations}
\begin{figure}[t]
    \centering
    \includegraphics[width=220pt]{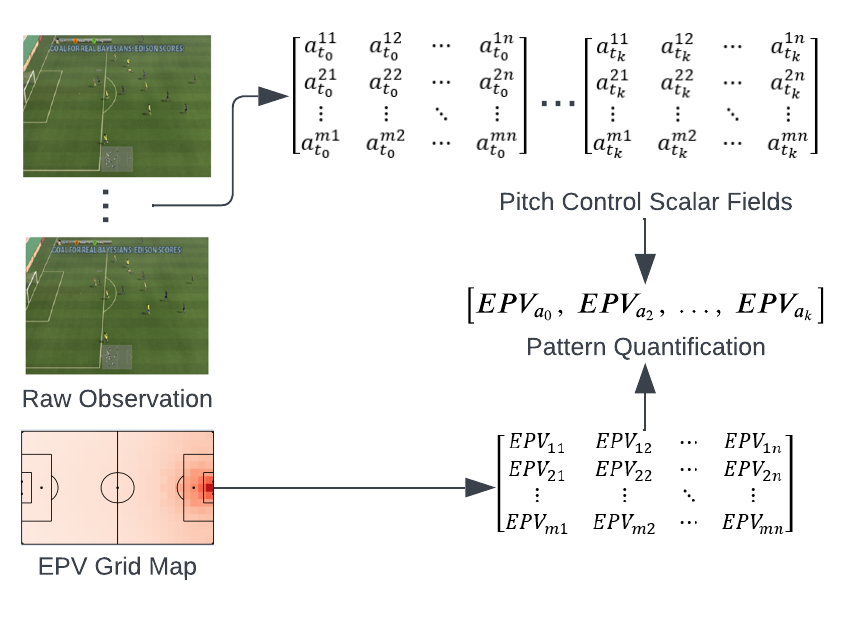}
    \caption{Quantification of pattern detected in raw observation}
    \label{quant}
\end{figure}

The raw observation at each timestep within learning contains all the contextual information in the current game state. As shown in Fig.~\ref{quant}, we generate the Pitch Control scalar field from the raw observation including players' and ball's positions and speed by utilizing the Pitch Control Function. In the Pitch Control scalar field, the probability of the attacking team controlling the ball at location $[m,n]$ at time \(t_k\) is represented by \(a_{t_k}^{mn}\). This Pitch Control scalar field extracts the team space-control pattern at the given game state. To evaluate and quantify this pattern, the EPV model is utilized. With the EPV model, the football pitch can be converted to an EPV grid map in which the expected possession value for ball at location $[m,n]$ is \({EPV}_{mn}\). In result, the expected possession value at \(t_k\) can be computed following Eq.~\eqref{epv}. \({EPV}_{a_k}\) (the sum of product of Pitch Control probability and EPV value across the whole pitch) denotes the EPV for game state at \(t_k\) showing how likely the game state will develop into a scoring chance. For a specific learning step, a lower game-state EPV suggests a better defending performance from agents. This is implemented into the reward function to let agents learn playing football from domain knowledge.

\begin{equation}
\label{epv}
{EPV}_{a_k}=\sum_{i=1}^{m}\sum_{j=1}^{n}{EPV}_{ij}\ast\ a_{t_k}^{ij}
\end{equation}

\subsection{Reward Shaping}
\begin{figure}[b]
    \centering
    \includegraphics[width=180pt]{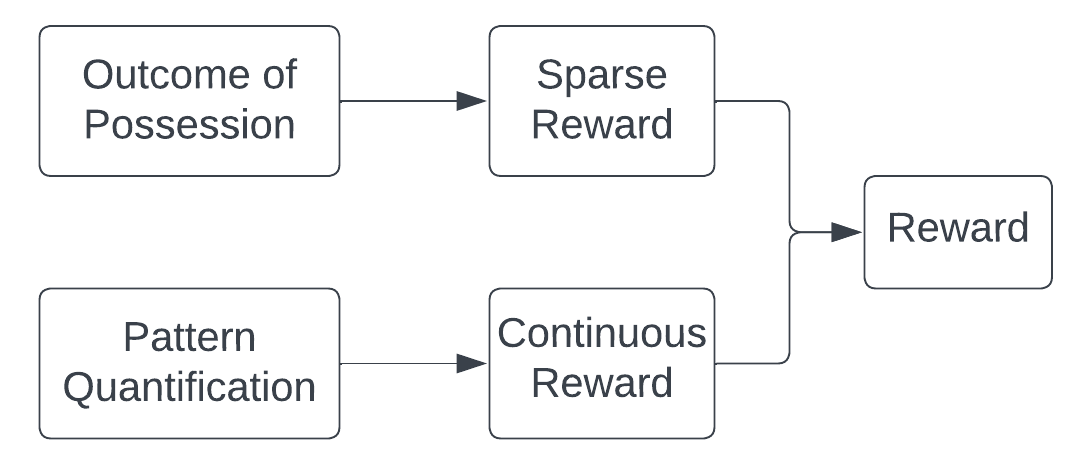}
    \caption{Reward shaping}
    \label{reward}
\end{figure}

To solve the reward sparsity problem while training agents we do reward shaping. As shown in Fig.~\ref{reward}, in addition to the primary sparse reward, outcome of each learning episode (goal difference between sides), we implement game-state EPV, the quantification of pattern detected, in the observation as continuous reward. At every learning step, each agent can receive immediate feedback about how negative or positive the impact of current game state on the long-term outcome of the episode.

\section{Experiments}

\subsection{Environment Setup}
To test our learning framework, we do the experiments in GRF, a simulated real football environment in which the players can be controlled by Reinforcement Learning algorithms and frameworks or built-in hard-coded AI. We set the game scenario as five players including one lazy goalkeeper staying at goal line without moving and four agents controlled by our framework, defending against 6 attacking players controlled by built-in AI. The reason why the goalkeeper is set as lazy is because we aim to train players excluding the goalkeeper to play defence, and the active goalkeeper will make the reward signal noisy. In specific, the agents can be free of penalty when the goalkeeper saves a good scoring chance. The reason why we set the number of attacking AI higher than number of controlled agents is that we want to make the game scenario more challenging for agents to learn the effective defending policy. Only open play game states are allowed, while the learning episode ends immediately if the ball goes out of bound, either side commits a foul or attacking side scores.
\subsection{Evaluation Metric}
To test the efficiency of algorithms or learning frameworks in the GRF environment, the goal difference is used as the evaluation metric \cite{kurach2020google}. It directly evaluates the trained agents' performance with game outcome. We use the goal difference as the evaluation metric and compare our learning framework against VDN without reward shaping. We firstly do quantitative experiments by comparing the learning curve and the average goal difference between two training frameworks. Then we make qualitative analysis by comparing the two frameworks in terms of the agents' performance in simulated football environment.
\subsection{Quantitative Results}
We train agents using VDN with our reward shaping method and VDN without reward shaping both for 2 million steps. For each learning framework, we run five random seeds in training. The difficulty of built-in attacking AI is set to be 0.95 (the highest difficulty). We compare their performances in two ways. We firstly track the episode returns in terms of the goal difference from agents trained by each framework. Secondly we test each group of trained agents for 32 episodes and compare the average goal difference per episode.

\begin{figure}[t]
    \centering
    \includegraphics[width=220pt]{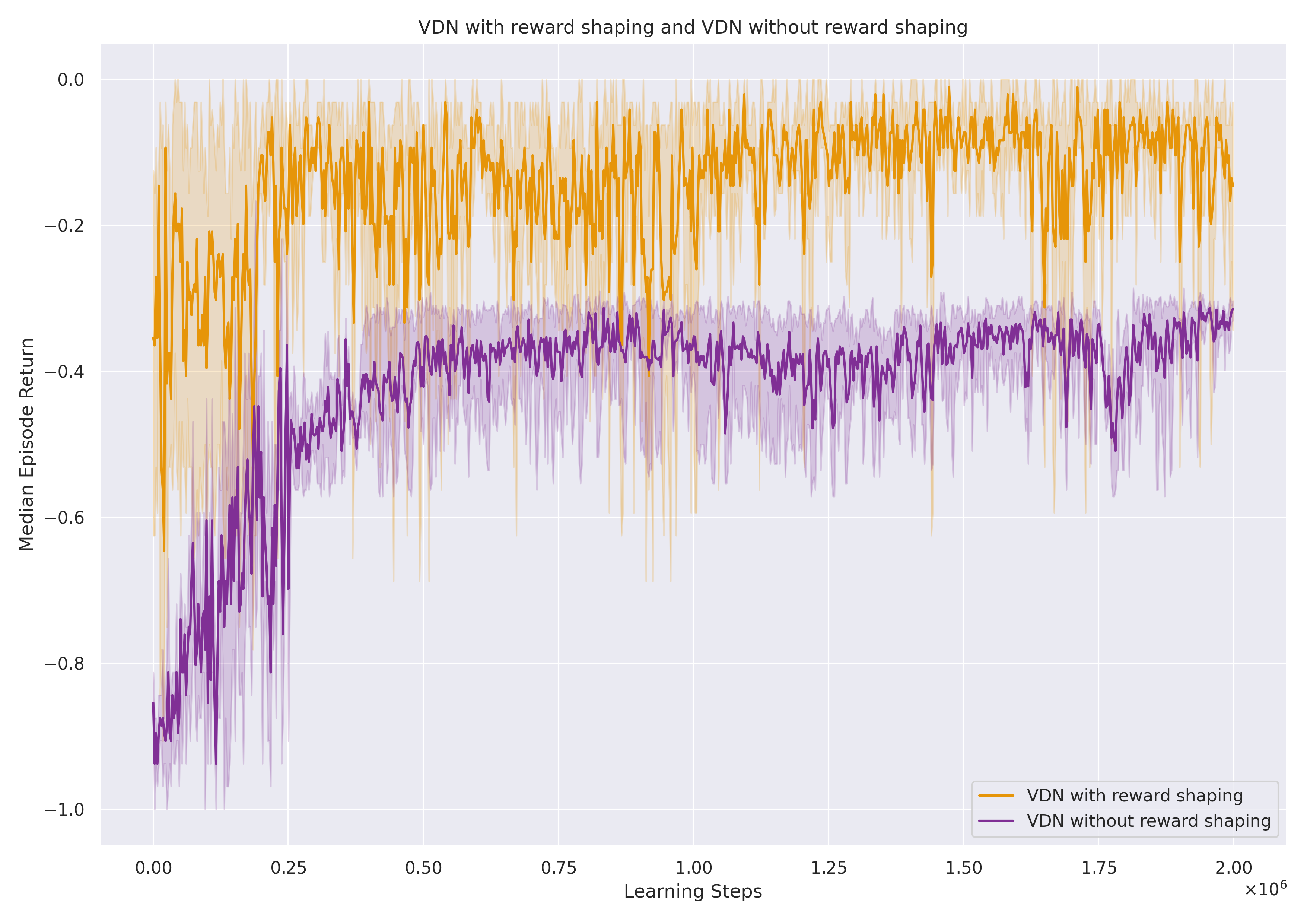}
    \caption{Median and quartile curves for goal difference of VDN with reward shaping and VDN without reward shaping}
    \label{learningcurve}
\end{figure}

We test the policy learnt by each framework for 32 episodes after every 2000 steps training. Fig.~\ref{learningcurve} shows the performance from each framework in terms of their median and quartile curve across 2 million steps training, ran with five independent random seeds. Our framework outperforms VDN according to the graph. From the yellow curve representing our learning framework and the purple curve denoting VDN, the episode return from our framework is higher. It indicates that the novel reward shaping enabled by pattern detection in observation and reward shaping through embedding contextual information help agents learn better when interacting with the complex environment. Moreover, as shown in the graph, our framework enables the agents to learn faster at the beginning, which can be encouraged by the domain knowledge embedded in the feedback to agents' actions. The computational time of our model is longer than VDN due to the calculation of EPV per timestep. However, we embed the contextual information in the reward signal at every timestep, which helps the model surpass VDN in terms of training intelligent agents to defend more efficiently in complex contexts.

Moreover, we run the last saving checkpoints from our framework and VDN on 32 new episodes with all three difficulties and calculate the average goal difference per episode to test the policy learnt by our framework. As shown in Table \ref{tab1}, agents controlled by our policy perform better than those controlled by VDN in every difficulty. Moreover, the difference between our framework and VDN in terms of goal difference achieved is bigger in hard environment than in the others. It indicates the advantage of our agents when competing against stronger opponents. Moreover, in easy and medium environments, the built-in attacking AI can miss lots of easy chances and the goal difference is not sensitive enough to judge the agents' performance precisely. In the environment with difficulty set as 0.95, only when the agents defend well in terms of preventing opponents easily getting close to the goal, the built-in AI will have lower scoring probability. As a result, the stability of our agents' performance across all three difficulties proves that our learning framework is a good addition to state-of-the-art MARL algorithms for training agents in complex environments like football.

\begin{table}[b]
\caption{Average Goal Difference Achieved by Policies Controlled by Our Framework and VDN}
\begin{center}
\begin{tabular}{|c|c|c|}
\hline
\textbf{Difficulty}&\multicolumn{2}{|c|}{\textbf{Frameworks Controlling Policies}} \\
\cline{2-3} 
\textbf & \textbf{\textit{VDN with Reward Shaping}}& \textbf{\textit{VDN}} \\
\hline
hard(0.95)& -0.15& -0.40 \\
medium(0.60)& -0.12& -0.23 \\
easy(0.05)& -0.07& -0.12 \\
\hline
\end{tabular}
\label{tab1}
\end{center}
\end{table}

\subsection{Qualitative Results}
To evaluate the policy trained by our learning framework more comprehensively, we do the qualitative experiment on policies learnt both by VDN with reward shaping and VDN without reward shaping. The trained agents are let play a new episode against built-in AI with difficulty set as hard and we visualize the game states within this episode to analyse their performance. We firstly analyse the performance from agents trained by VDN. Secondly we evaluate the performance from agents trained by our learning framework and discuss the reasons why VDN with reward shaping is believed to be more effective when compared to VDN without reward shaping to learning policy of defending in Google Football.
\begin{figure*}
    \centering
    \includegraphics[width=\textwidth]{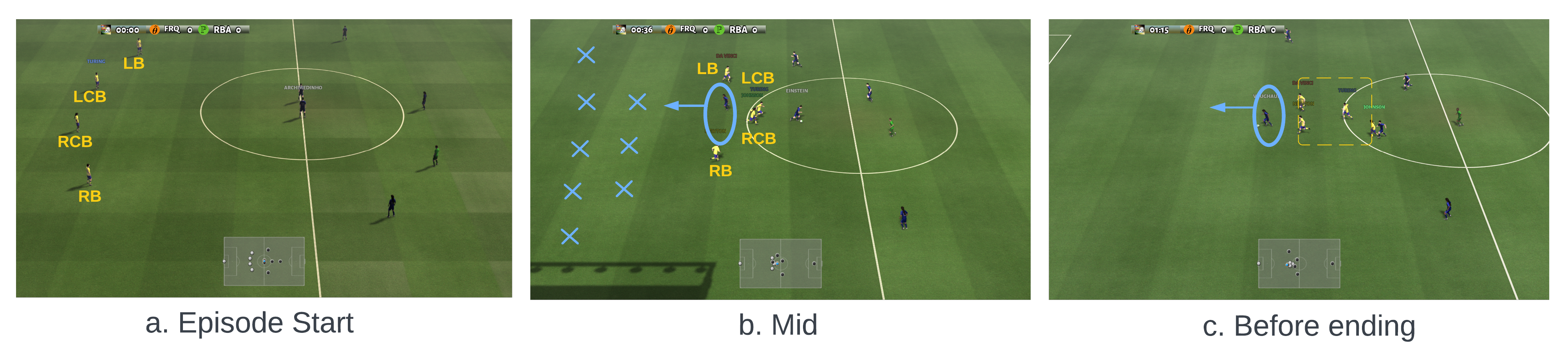}
    \caption{Performance Visualization from agents trained by VDN}
    \label{vis1}
\end{figure*}
\begin{figure*}
    \centering
    \includegraphics[width=\textwidth]{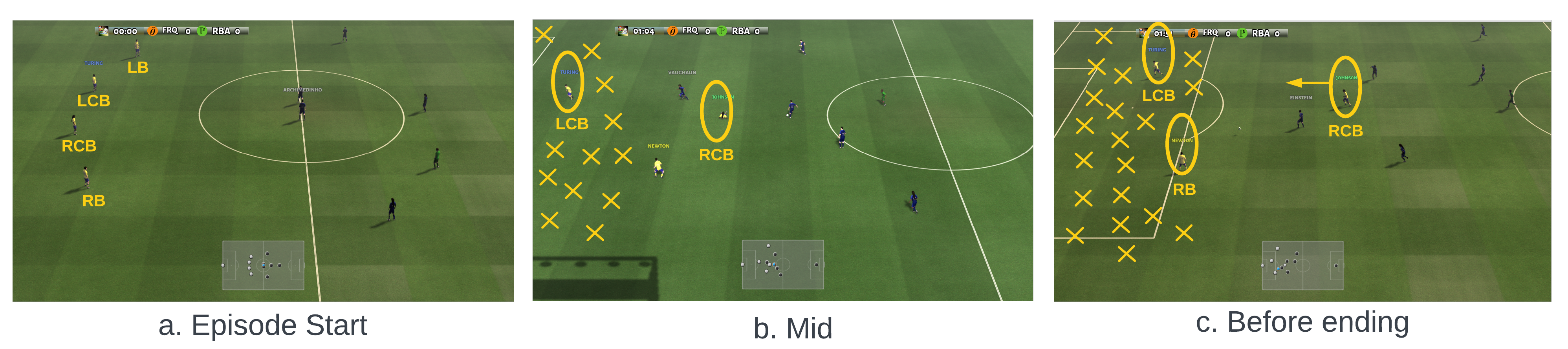}
    \caption{Performance Visualization from agents trained by our learning framework}
    \label{vis2}
\end{figure*}
As shown in Fig.~\ref{vis1}, the three frames from left to right demonstrate three stages of the given episode from the starting point to the state before the end. The episode starts when the blue team controlled by the environment built-in AI kicks off in the halfway line. The yellow team controlled by VDN spread in the defensive half ready for preventing opponents from scoring the goal. In the second stage of the episode, as shown in frame b, all the four agents push high towards the opponents trying to tackle the ball from ball carrier. However, since none of the defenders stays behind, attacking side is given large open space in front to utilize, marked with blue crosses \cite{fernandez2018wide}. The off-ball movement from the opponent team doesn't draw the attention from agents controlled by VDN, which leads to the third stage where the off-ball runner from attacking side easily receives the pass from the original ball carrier and is able to attacking the goal without any barrier \cite{herold2022off}. The cause of the failure from agents can be the credit assignment problem when using VDN to learn the policy. At first, when the agents learn through VDN without reward shaping, the only reward signal they get when interacting with the environment is the goal difference. In specific, when the opponent team scores, they receive a negative reward. The sparse reward signal limits their understanding to the environment and they cannot receive feedback for the impact of every action on the final outcome. Running towards the ball carrier and making attempts to tackle the ball is learnt to be one of the key policies to stop the attacking side from scoring. However, the simulated environment, just like a real football game, contains too many contextual factors deciding the outcome of each possession. This simple defensive strategy shown in Fig.~\ref{vis1} can only be profitable in terms of winning the ball back if the attacking team forces the pass making turnover while in many other scenarios the agents easily concede the goal for allowing the off-ball runner to utilize the huge open space on the pitch.
For testing the performance from agents trained by our learning framework, the difficulty is set to 0.95, the same as the above episode. Fig.~\ref{vis2} demonstrates the three stages in the visualized testing episode. After kick-off, instead of pushing directly towards the opponent, the defending team controlled by our policy chooses to stay compact controlling the area in front of the goal. According to frame in Fig.~\ref{vis2}b, one of our agents, the right center-back (RCB), defends aggressively by moving towards the ball carrier intending to make challenges while the other center-back (LCB) stays behind to protect the space behind RCB instead of pushing together. The cooperation from these two center-backs enables the compact defensive formation and at the same time prevents the opponents to easily attack into the final third of the pitch \cite{shaw2019dynamic}. In the last stage, the opponent makes the pass to the final third of the defensive side. LCB and RCB quickly move to the edge of penalty box and stay close to each other forming the defensive line to stop opponents making direct threats to the goal. In the meantime, RCB chases back to give the potential pass receiver pressure after failing to intercept pass. This episode ends as a success for the defensive side when RB (Right-back) intercepts the pass. In all stages of this episode, our agents are consistently aware of covering the in-front-of-goal space. This space is the most crucial area for both defensive and attacking sides \cite{low2021porous}. Using Pitch Control Function to detect the pattern from observation and embedding contextual information by EPV makes the shaped reward an informative reflection of the game state. At every step within the episode, our agents can receive the reward signal indicating whether their actions make positive or negative impact. Moreover, actions from RCB within this demonstrated episode can be encouraged by two factors. At first, since the most valuable area has already been controlled by the other teammates, RCB's presence in that area will not increase the reward. As a result, RCB moves forward aiming to control the area among attacking players and increase the reward value. Secondly, the sparse reward, goal difference can be the other factor in deciding RCB's actions. In specific, if the ball carrier from the attacking team receives no pressure, it will be more likely for him to make threats to the goal. RCB's pressure can reduce the opponent's scoring probability and therefore avoid the whole team receiving the negative reward, which can be equivalent to policy learnt by VDN without reward shaping. In general, the reward shaping allows our learning framework to outperform VDN in learning football defending policies in Google Football.
\raggedbottom\section{Conclusion}
In this work, we proposed a novel reward shaping method to train multiple agents in complex environment. We do the research in the Google Football environment by training agents to play defence through our learning framework in which we modify the primary reward, goal difference, by embedding the contextual information, expected possession value from game state observation. As shown in the experiment results, our reward shaping method increases the learning efficiency of VDN in environment with sparse reward signal. The agents receive step-by-step credit assignment enabled by state-of-the-art football performance analysis model in the learning process, which makes them have more stable performance when facing different situations. Moreover, the proposed method of training agents in Google Football can be applied in other areas. In environments involving multiple agents and having sparse reward signal, like football games, the learning efficiency of MARL algorithms can be improved by embedding the contextual information in a shaped reward. Specifically, for a given environment, domain knowledge from real-life related area can be used to quantify the game state observations to continuous reward signals which can help reach a better learning outcome. The main limitation of this research is that the learning framework is only tested in a simplified version of original environment with less agents and only defending scenarios involved. The future research will focus on developing a learning framework which can be applied on more complex environments.   

\bibliographystyle{IEEEtranN}
\bibliography{references}

\end{document}